\documentclass[review]{elsarticle}

\usepackage{lineno,hyperref}
\usepackage{comment}
\usepackage{amsmath}
\usepackage{amsthm}
\usepackage{amsfonts}
\usepackage{algorithm}
\usepackage{graphicx}
\usepackage{subfigure}
\usepackage[dvipsnames]{xcolor}
\usepackage[noend]{algpseudocode}
\hypersetup{pdfauthor={Name}}

\newcommand{\red}[1]{\textcolor{red}{{#1}}}

\newcommand{\orange}[1]{\textcolor{orange}{{#1}}}
\newcommand{\green}[1]{\textcolor{ForestGreen}{{#1}}}

\modulolinenumbers[5]

\journal{Knowledge-Based Systems}

\begin{document}

\begin{frontmatter}

\title{{A Condense-then-Select Strategy} \\{for Text Summarization}}

\author[Macau_U]{Hou Pong Chan\corref{mycorrespondingauthor}}
\ead{hpchan@um.edu.mo}

\author[CUHK_CSE]{Irwin King}
\ead{king@cse.cuhk.edu.hk}

\address[CUHK_CSE]{Department of Computer Science and Engineering, The Chinese University of Hong Kong, Shatin, N.T., Hong Kong SAR, China.}
\address[Macau_U]{Department of Computer and Information Science, University of Macau, Macau SAR, China.}

\cortext[mycorrespondingauthor]{Corresponding author}

\begin{abstract}
Select-then-compress is a popular {hybrid} framework for {text summarization} due to its high efficiency. This framework first selects salient sentences and then independently condenses each of the selected sentences into a concise version. However, compressing sentences separately ignores the context information of the document, and is therefore prone to delete salient information. To address this limitation, we propose a novel \emph{condense-then-select} framework for {text summarization}. Our framework first concurrently condenses each document sentence. Original document sentences and their compressed versions then become the candidates for extraction. Finally, an extractor utilizes the context information of the document to select candidates and assembles them into a summary. If salient information is deleted during condensing, the extractor can select an original sentence to retain the information. Thus, our framework helps to avoid the loss of salient information, while preserving the high efficiency of sentence-level compression. Experiment results\footnote{Code will be available at \url{https://github.com/kenchan0226/abs-then-ext-public}} on the CNN/DailyMail, DUC-2002, and {Pubmed} datasets demonstrate that our framework outperforms the select-then-compress framework and other strong baselines. 
\end{abstract}


\end{frontmatter}


\section{Introduction}
Text summarization aims at distilling the core information of a document into a short and concise summary. 
Existing summarization methods can be divided into three categories: \emph{extractive}, \emph{abstractive}, and \emph{hybrid}. Extractive methods select salient sentences from the input document to form a summary prediction. 
Abstractive methods, on the other hand, generate a summary word by word from scratch, and can introduce novel words that do not appear in the document.
Though recent abstractive methods~\cite{DBLP:conf/conll/NallapatiZSGX16,DBLP:conf/acl/SeeLM17,DBLP:conf/iclr/PaulusXS18,DBLP:conf/acl/keyphraseACL20} have achieved promising results by adopting the encoder-decoder neural model~\cite{DBLP:conf/iclr/BahdanauCB14}, they suffer from the problem of slow training and decoding.

\begin{figure}[t!]
\centering
\includegraphics[width=0.75\columnwidth]{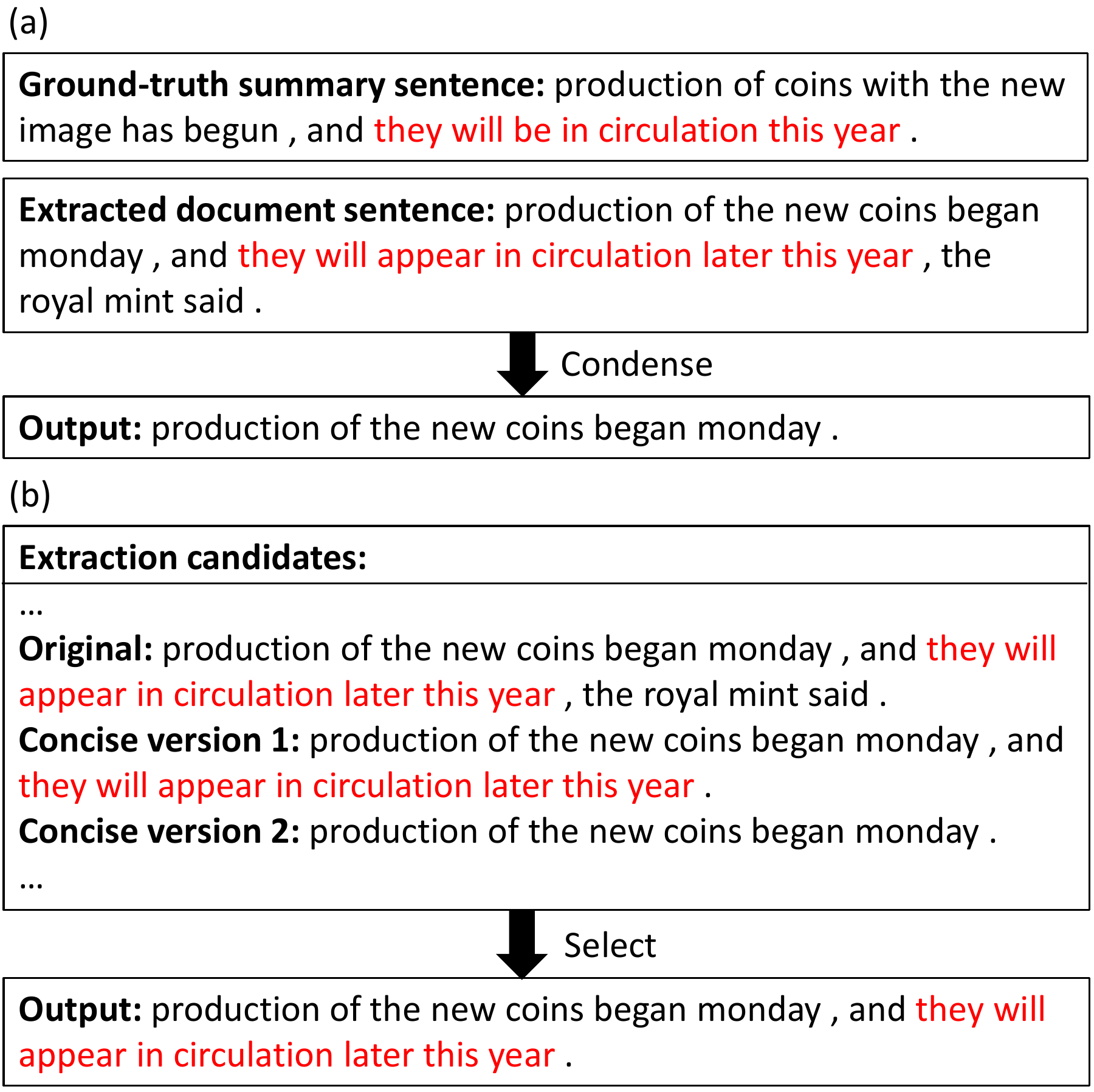}
\caption{
(a) A sample ground-truth summary sentence and a document sentence extracted by the sentence rewriting method~\cite{DBLP:conf/acl/BansalC18}. The extracted sentence is then condensed into an output sentence. 
The omitted salient information is highlighted. 
(b) Sample extraction candidates produced by our framework. Our compression-controllable abstractor condenses a document sentence into two concise versions. An extractor then selects ``concise version 1" to produce an output sentence.
}
\label{figure:intro-example}
\end{figure}

Hybrid methods combine extractive and abstractive methods by adopting a select-then-compress framework. These methods first select salient sentences from the input document, and then condense each of the selected sentences \mbox{independently} into a more concise version. Different hybrid methods apply different models~\cite{DBLP:conf/aaai/KnightM00,DBLP:conf/anlp/JingM00,DBLP:conf/acl/0001L16} to condense sentences. 
A recent sentence rewriting method~\cite{DBLP:conf/acl/BansalC18} adopts the encoder-decoder neural model as the condensing model, which achieves promising results. Compared to abstractive methods, hybrid methods are easier to train as a condensing model only needs to condense one document sentence instead of the whole input document. Moreover, the decoding of each summary sentence can be done in parallel. Hence, hybrid methods have substantially higher training and testing efficiency. 

However, since hybrid methods condense each sentence separately, they cannot utilize the context information of the input document to determine which information to discard during condensing. Thus, hybrid methods are prone to delete salient information from a document sentence, leading to a loss of salient information in the output summary. 
For example, Figure~\ref{figure:intro-example}~(a) shows a document sentence being selected by the sentence rewriting method~\cite{DBLP:conf/acl/BansalC18}. The sentence is then condensed into a concise version that appears in the output summary. 
We observe that the salient information of ``will appear in circulation later this year'' was deleted by the framework, while such information appears in the ground-truth summary sentence. 

To address the problem of salient information deletion while preserving the fast and parallel decoding of sentence-level compression, we propose a novel \emph{condense-then-select} summarization framework as follows. An \emph{abstractor} first concurrently condenses each input document sentence into several versions, e.g., with varying compression ratios. The original sentences and their compressed versions then become the candidates for extraction. 
Next, an \emph{extractor} makes a sequence of decisions, each time selecting either one original sentence or its best condensed version, conditioned on the selection history and the context of the document. 

We show an example in Figure~\ref{figure:intro-example}~(b), where our framework uses an abstractor to generate two compressed versions of the original sentence. The extractor selects the concise version that covers the important information on circulation and also avoids the inclusion of minor content (``the royal mint said''). As a result, the output summary contains more salient information than the summary generated by a select-then-condense framework. 
In the extreme case where salient information is absent in all condensed sentences, the extractor in our framework can select original sentences to retain the salient information. 
Our framework therefore benefits from (1) fast training and testing of sentence-level abstraction, (2) conciseness of abstractive method, and (3) extraction candidates with both original and condensed sentences to avoid the loss of salient information. 

In this work, we explore different types of abstractors to condense a document sentence into different compressed versions. Among these, the most effective one is found to be our proposed \emph{compression-controllable abstractor}, which generates two concise versions that retain the information of a sentence to various extents. The output from this abstractor is illustrated in Figure~\ref{figure:intro-example}, where version 2 has a higher compression ratio. 

We conduct extensive empirical studies to evaluate our condense-then-select framework. Experiment results on the CNN/DailyMail dataset show that our framework outperforms the select-then-compress framework and other strong baselines. 
Moreover, we analyze how different types of abstractors affect the overall performance of our framework. 
Furthermore, experimental results on the DUC-2002 dataset demonstrate that our framework achieves better generalization than the baselines. 
{Finally, experiments on the Pubmed dataset show that our framework outperforms strong long text summarization baselines. }

Our contributions can be summarized as follows: (1) a new condense-then-select framework for {text summarization}; (2) an extensive analysis of different types of abstractors within our framework; and (3) our framework outperforms strong baselines on both CNN/DailyMail and DUC-2002 datasets.


\section{Related Work}
\smallskip
\noindent \textbf{Extractive methods. }
Most of the previous extractive methods consist of two major components: scoring of each document sentence and selection of sentences. 
Sequence labelling models~\cite{DBLP:journals/corr/NallapatiZM16,DBLP:conf/aaai/NallapatiZZ17,DBLP:conf/naacl/NarayanCL18,DBLP:conf/aaai/WuH18} first learn a relevance score for each document sentence, and then select the document sentences with a score larger than 0.5. The joint scoring and selection approach~\cite{DBLP:conf/acl/ZhaoZWYHZ18,DBLP:conf/acl/BansalC18} is proposed to extract sentences one by one. At each step, a decoder attends to the document sentences and selects the sentence with the highest attention score. 
Some of the methods above~\cite{DBLP:conf/naacl/NarayanCL18,DBLP:conf/aaai/WuH18,DBLP:conf/emnlp/DongSCHC18,DBLP:conf/acl/BansalC18} apply reinforcement learning (RL) to optimize the non-differential ROUGE scores. 
{Recently, Zhong et al.~\cite{DBLP:conf/acl/TextMatching20} propose a method to perform extraction at summary level instead of sentence level. This method first enumerates all possible combinations of document sentences as the summary candidates from a pruned document. Then it selects the candidate that has the highest semantic similarity with the document measured by a text matching model. } 
Compared to the above extractive methods, our framework uses an abstractive model to condense the document sentences before the sentence extraction. Hence, our framework can introduce novel words in the output summary. 

\smallskip
\noindent \textbf{Abstractive methods. }
Abstractive methods rely on the encoder-decoder neural model~\cite{DBLP:conf/iclr/BahdanauCB14} to generate a summary. 
See et al.~\cite{DBLP:conf/acl/SeeLM17} propose the pointer-generator network that allows the decoder to copy words from the input document. 
Gehrmann et al.~\cite{DBLP:conf/emnlp/GehrmannDR18} introduce the bottom-up attention to improve the contention selection ability of the pointer-generator network, whereas other methods~\cite{DBLP:conf/acl/SunHLLMT18,DBLP:conf/naacl/KeyphraseNAACL19,DBLP:conf/sigir/ReviewSumSIGIR20} apply multi-task learning frameworks to enhance the contention selection capability. 
{The ASGARD method~\cite{DBLP:conf/acl/KGSum20} incorporates the information from a knowledge graph into the decoder to improve the faithfulness of the generated summary. }
Several methods~\cite{DBLP:conf/iclr/PaulusXS18,DBLP:conf/naacl/CelikyilmazBHC18} apply RL to directly optimize the ROUGE scores. 
Since the training objective of ROUGE scores harms the language fluency of a decoder~\cite{DBLP:conf/acl/BansalC18}, we only use RL to train our extractor in order to improve ROUGE scores without hurting the fluency. 
{Recent abstractive methods~\cite{DBLP:conf/acl/BART20, DBLP:conf/emnlp/BertSum19, DBLP:journals/corr/UniLMv2_2020, DBLP:conf/icml/PEGASUS}} fine-tune large pre-trained language models on text summarization datasets and achieve state-of-the-art performance. 
{Xu et al.~\cite{DBLP:conf/acl/SelfAttnGuidedCopy20} uses the self-attention distribution in the Transformer encoder to estimate the centrality of each source word. Then the centrality of each word is used to guide the copying process in a large pre-trained language model. }
Another line of research~\cite{DBLP:conf/aclnmt/FanGA18,DBLP:conf/emnlp/LiuLZ18} studies the problem of controlling the absolute length of the output summary, whereas our compression-controllable abstractor only controls the length difference between the input and output sentences. 
Compared to these methods, our framework only performs sentence-level abstraction, which has higher training efficiency. 

\smallskip
\noindent \textbf{Hybrid methods. }
Existing hybrid methods follow a select-then-compress framework that first selects document sentences and then condenses each of them. Traditional methods adopt integer linear programming~\cite{DBLP:journals/coling/ClarkeL10}, Hidden Markov Models~\cite{DBLP:conf/anlp/JingM00}, and statistical models based on a parsing tree~\cite{DBLP:conf/aaai/KnightM00}. 
Some recent methods use neural networks to compress a selected document sentence by extracting important words~\cite{DBLP:conf/acl/0001L16} or selecting a compression option derived from constituency parses~\cite{syntactic_compress_19}, but they cannot generate novel words. 
The sentence rewriting method~\cite{DBLP:conf/acl/BansalC18} applies the encoder-decoder model to rewrite the selected document sentences into shortened versions, which can generate novel words. 
The SENECA model~\cite{DBLP:conf/emnlp/SENECA19} explicitly incorporates the entity information into the extractor module. 
However, sentence-level compression is prone to delete salient information. We address this problem by adopting a condense-then-select strategy, which shares a similar idea with the trimming approach~\cite{zajic2006sentence} for multi-document summarization. The trimming approach first compresses each leading document sentence into multiple candidates by heuristic rules, and then selects sentences from the candidates. We are the first to study this idea in abstractive summarization, where our abstractor can generate novel words. 

{
Although our work is closely related to the sentence-rewriting method~\cite{DBLP:conf/acl/BansalC18}, there are four key differences. 
(1) The sentence-rewriting method follows a select-then-compress framework; whereas our methods follow a condense-then-select framework. If salient information is deleted during content abstraction, the extractor in our framework can select original sentences to retain the salient information. 
(2) In the sentence-rewriting method, the extraction candidates only contain the original document sentences. Hence, the candidate extractor only considers the interactions among the document sentences. In our framework, the extraction candidates have two types of interaction: the interaction among different document sentences; and the interaction between a document sentence and its condensed versions. Hence, we add a sentence candidate set representation into the candidate extractor to model the interaction between a document sentence and its condensed versions. 
(3) The sentence-rewriting method uses the pointer-generator network~\cite{DBLP:conf/acl/SeeLM17} as the abstractor. Our work extends the pointer-generator network to build a compression-controllable abstractor, which condenses a document sentence into two concise versions with different compression levels. 
(4) During training, the sentence rewriting method~\cite{DBLP:conf/acl/BansalC18} uses the sentence-level Rouge-L score as the reward of selecting a candidate, which aims at a local optimum of one selection. Our work uses the marginal increase in ROUGE-L score of the output summary as the reward, which aims at the global optimum of the entire output summary. 
}

\section{Condense-then-Select Framework}
We define the problem of text summarization as follows. Given a document $\mathbf{X}$ with a sequence of $n$ sentences, $\mathbf{X}=(\mathbf{x}_{1},\mathbf{x}_{2},\ldots,\mathbf{x}_{n})$, the output is a summary containing a sequence of $m$ sentences, $\mathbf{Y}=(\mathbf{y}_{1},\ldots,\mathbf{y}_{m})$. 

\begin{figure}[t]
        \centering
        \includegraphics[width=0.8\columnwidth]{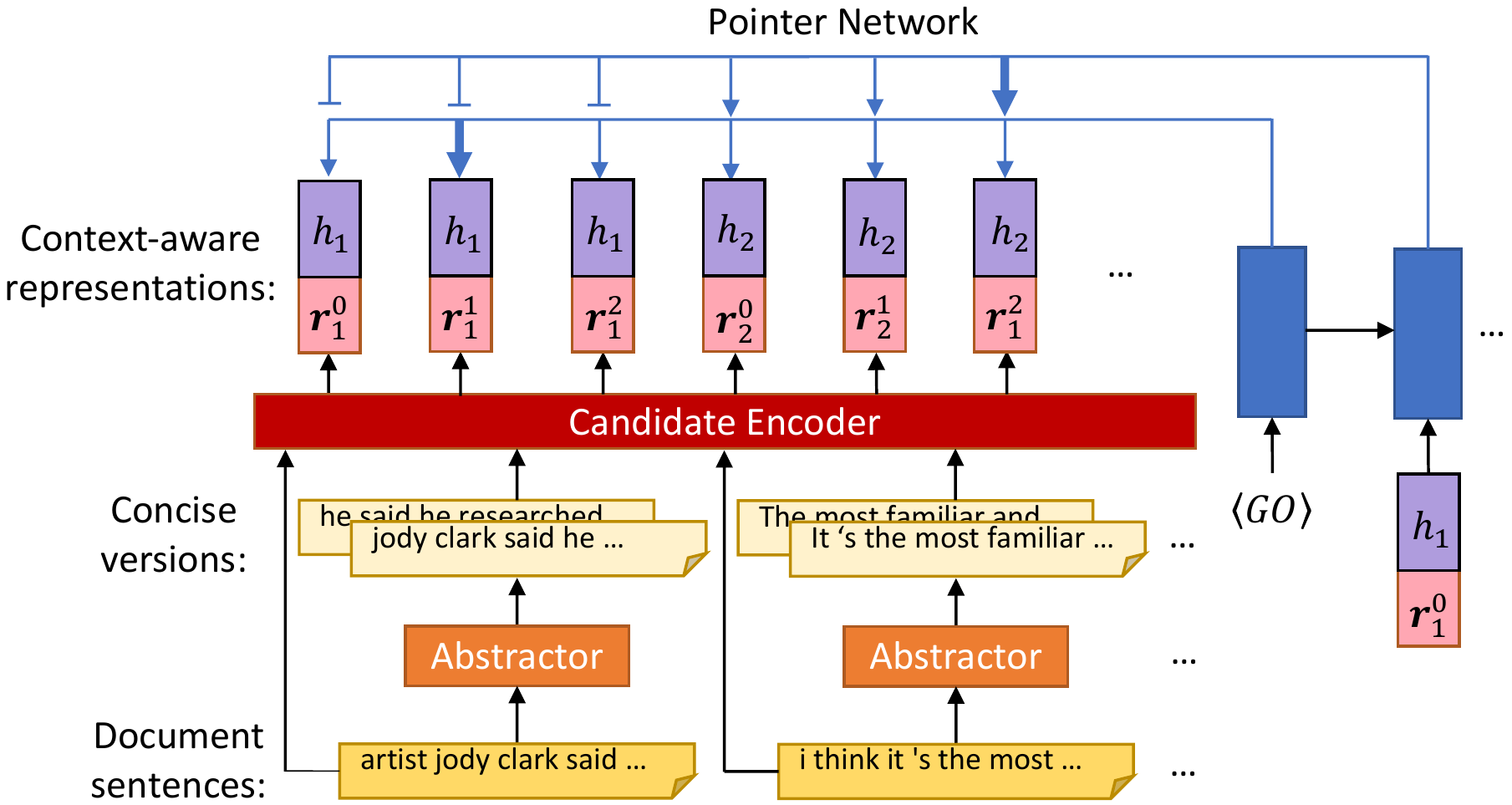}
        \caption{Overall framework architecture. An abstractor first condenses each document sentence into different concise versions. Next, a candidate encoder converts all extraction candidates into context-aware representations. Finally, a pointer network extracts candidates based on their representations. $\langle GO \rangle$ denotes a trainable initial input. }
        \label{fig:overall-architecture}
        \vspace{-0.1in}
\end{figure}

Our condense-then-select framework is comprised of an abstractor and an extractor. 
First, the abstractor condenses each document sentence $\mathbf{x}_{i}$ into $k$ different concise versions $\mathbf{x}_{i}^{1}, \ldots, \mathbf{x}_{i}^{k}$. 
All document sentences $\mathbf{x}_{i}$ and their concise versions $\mathbf{x}_{i}^{j}$ then become the candidates for extraction. 
After that, the extractor applies a candidate encoder to learn a context-aware representation for each candidate and uses a pointer network~\cite{DBLP:conf/nips/VinyalsFJ15} to select sentence candidates and produce a summary prediction $\hat{\mathbf{Y}}$. The overall architecture is illustrated in Figure~\ref{fig:overall-architecture}. 

\subsection{Abstractor Modules}\label{sec:abstractor}
In this work, different types of abstractors are explored. 
All of our abstractors rely on the pointer-generator network~\cite{DBLP:conf/acl/SeeLM17} to learn a mapping between an input sequence and an output sequence. Specifically, the input sequence is a document sentence, and the output sequence is a concise version of it. 
We use the maximum likelihood loss to train this model with training pairs of a document sentence and its concise version. Different abstractors vary in how such training pairs are constructed. They are detailed as follows. 

\smallskip
\noindent \textbf{One-to-one top-$\mathbf{k}$ abstractor. }
This abstractor uses the method proposed by Chen and Bansal~\cite{DBLP:conf/acl/BansalC18} to align each summary sentence $\mathbf{y}_{t}$ with a document sentence $\mathbf{x}_{j_{t}}$. It then treats the pairs of $(\mathbf{x}_{j_{t}}, \mathbf{y}_{t})$ as the training pairs for the pointer-generator model.
The alignment rule is as follows:
for each summary sentence $\mathbf{y}_{t}$, we find a document sentence $\mathbf{x}_{j_{t}}$ that covers as much information of $\mathbf{y}_{t}$ as possible:
%
\begin{align}\label{eq:primary-sentence}
    j_{t}=\text{argmax}_{i} \text{ Rouge-L}_{recall}(\mathbf{x}_{i}, \mathbf{y}_{t}) \text{.}
\end{align}
We refer to $\mathbf{x}_{j_{t}}$ as the \emph{source sentence} of $\mathbf{y}_{t}$. 
To generate $k$ different concise versions given a document sentence $\mathbf{x}_{i}$ during inference time, we use a beam search algorithm with a beam size of $N$.
The top-$k$ output sequences in the list then become the $k$ concise versions of a document sentence, $\mathbf{x}_{i}^{1}, \ldots, \mathbf{x}_{i}^{k}$. 

\smallskip
\noindent \textbf{One-to-one long-short abstractor. }
This abstractor shares the same model and training procedure with the one-to-one top-$k$ abstractor. The only difference is that it takes the longest and shortest sequences from the $N$-best list of beam search to be the two concise versions, $\mathbf{x}_{i}^{1}$ and $\mathbf{x}_{i}^{2}$, for an input document sentence $\mathbf{x}_{i}$.

\smallskip
\noindent \textbf{Compression-controllable abstractor. }
Our compression-controllable abstractor generates two concise versions of different compression levels\footnote{{We tried training an abstractor that generates concise versions of three different compression levels. However, we found that in many of the output samples, two of the three concise versions are the same as each other. Hence, we thought that it is difficult to condense an original document sentence into three different concise versions and we used two concise versions in our work.} }. Overall speaking, this abstractor maintains two embedding vectors to indicate two compression levels (high and low). Then at each decoding step, we feed the concatenation of the compression level embedding and the embedding of the previous predicted token to the decoder, which allows the generation of a concise version to be conditioned on the given compression level.

{
The followings are the details of our compression-controllable abstractor. We first construct a compression level label $\eta_{t}$ for each training pair of $(\mathbf{x}_{j_{t}}, \mathbf{y}_{t})$. 
Given a compression level $\eta_{t}$, this abstractor first maps $\eta_{t}$ to its corresponding compression level embedding, $\mathbf{\beta}_{t}\in \mathbb{R}^{d_{\beta}}$ via a look-up table. Then it uses the concatenation of $\mathbf{\beta}_{t}$ and the embedding of the previous predicted token to be the input of the decoder at every decoding step, i.e., $s_{t}=LSTM_{1}(s_{t-1}, [\mathbf{e}_{t-1};\beta])$, where $LSTM_{1}$ denotes the decoder LSTM used in the abstractor, $s_{t}$ denotes its $t$-th hidden state, $\mathbf{e}_{t-1}$ denotes the embedding of the previous predicted token. 
}

{
We define two compression levels, \emph{high} and \emph{low}, using the following rule: $(\mathbf{x}_{j_{t}}, \mathbf{y}_{t})$ has a high compression level if its compression ratio is larger than 0.5, otherwise, it has a low compression level. The compression ratio between a summary sentence $\mathbf{y}_{t}$ and its source sentence $\mathbf{x}_{j_{t}}$ is defined as $\frac{|\mathbf{x}_{j_t}| - |\mathbf{y}_{t}|}{|\mathbf{x}_{j_t}|}$, where $|\mathbf{x}|$ denote the number of tokens in $\mathbf{x}$. During training of the abstractor, we feed the embedding of the compression level associated with the reference summary sentence and its source sentence. 
}

{
After training, we use this abstractor to generate two concise versions for each document sentence $\mathbf{x}_{i}$ as follows. We first generate a concise version of low compression level by feeding the embedding of low compression level. Then we generate a concise version of high compression level by feeding the embedding of the high compression level\footnote{{Note that each compression level indicates a range of compression ratios, the exact compression ratio of a concise version is determined by the model.}}. 
In this abstractor, we pick the top output sequence from the beam search. 
}


\smallskip
\noindent \textbf{Two-to-one abstractor. }
We introduce a two-to-one abstractor that models the two-to-one alignment between two document sentences and one summary sentence. 
In addition to a source sentence $\mathbf{x}_{j_{t}}$, we use another document sentence $\mathbf{x}_{j'_{t}}$ to provide auxiliary information for the generation of $\mathbf{y}_{t}$. We refer to $\mathbf{x}_{j'_{t}}$ as the \emph{secondary source sentence} of $\mathbf{y}_{t}$. To construct the training tuples of $(\mathbf{x}_{j_{t}}, \mathbf{x}_{j_{t}'}, \mathbf{y}_{t})$, we use the following rule to find the secondary source sentence: 
%
\begin{align}\label{eq:secondary-sentence}
    j'_{t}=\text{argmax}_{i\neq j_{t}} \text{ Rouge-L}_{recall}([\mathbf{x}_{j_{t}} ; \mathbf{x}_{i} ], \mathbf{y}_{t}) \text{,}
\end{align}
where $[\cdot ; \cdot ]$ denotes concatenation. During training, we concatenate $\mathbf{x}_{j_{t}}$ and $\mathbf{x}_{j_{t}'}$ to form an input sequence. 
During inference, we treat the input document sentence $\mathbf{x}_{i}$ as the source sentence. 
Since we do not know which document sentence can provide auxiliary information during inference, we assume that neighboring document sentences have a higher chance of talking about the same topic. Thus, we treat each of its neighboring document sentences, $\mathbf{x}_{i-1}$ and $ \mathbf{x}_{i+1}$, as the secondary source sentence respectively. 
More specifically, we feed $[\mathbf{x}_{i}; \mathbf{x}_{i-1}]$ to the pointer-generator network to generate one concise version $\mathbf{x}_{i}^{1}$, then feed $[\mathbf{x}_{i}; \mathbf{x}_{i+1}]$ to the model to generate another concise version $\mathbf{x}_{i}^{2}$. 

\subsection{Extractor Module}\label{sec:extractor}
The extractor module first encodes each extraction candidate into a context-aware representation, and then uses a pointer network~\cite{DBLP:conf/nips/VinyalsFJ15} to extract candidate sentences sequentially. 

\begin{figure}[t]
        \centering
        \includegraphics[width=0.6\textwidth]{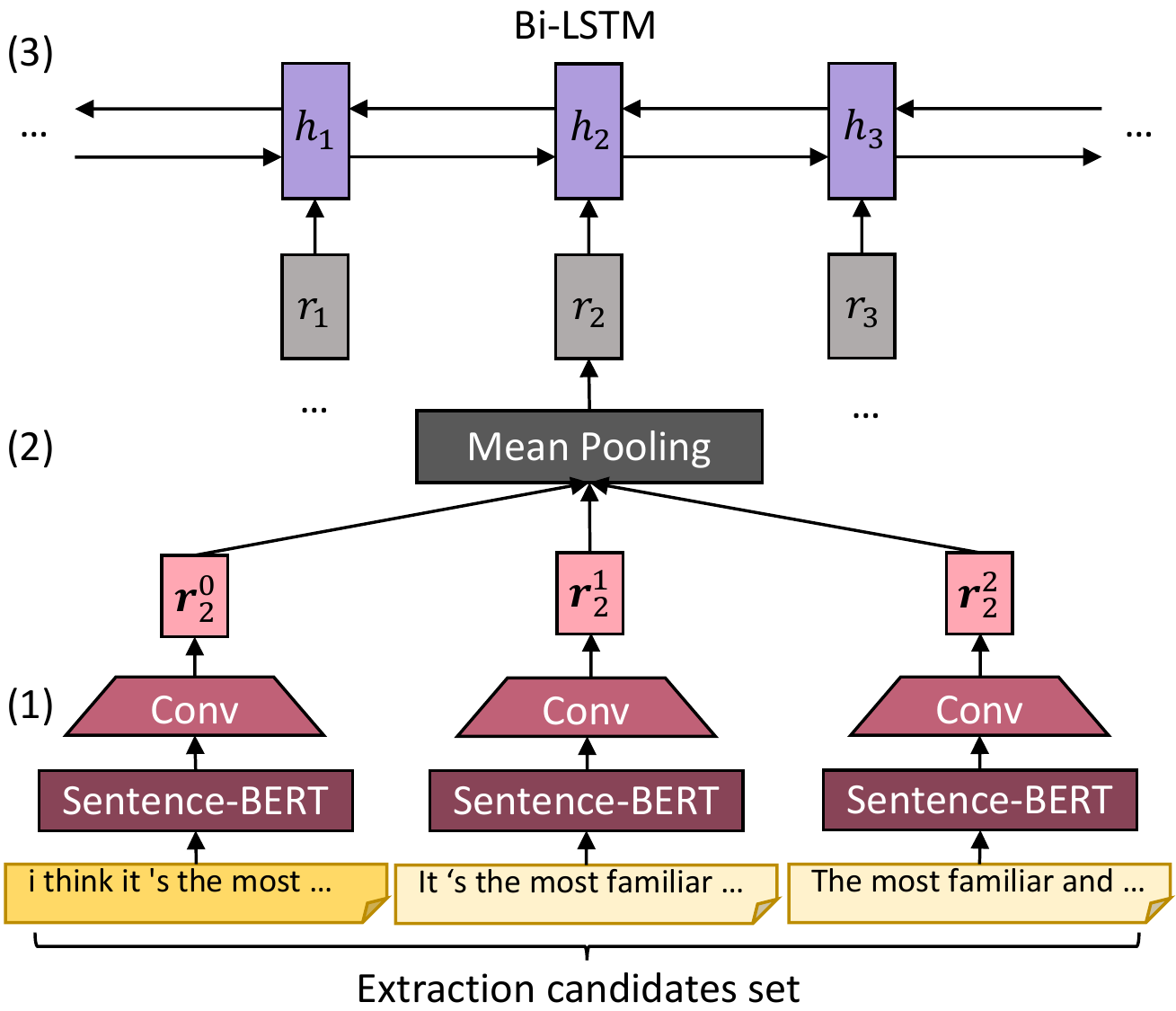}
        \caption{Architecture of the candidate encoder. (1) Sentence-BERT~\cite{DBLP:conf/emnlp/Sentence-BERT19} and CNN are applied to encode each candidate in the extraction candidate set. (2) Mean pooling is used to aggregate local candidate representations into a candidate set representation. 
        (3) All candidate set representations are fed through a Bi-LSTM to learn a document-level representation.}
        \label{fig:cand-encoder}
        \vspace{-0.1in}
\end{figure}

\smallskip
\noindent \textbf{Sentence candidate encoding. }
The sentence candidate encoder incorporates the document context in a hierarchical manner. 
For each document sentence $\mathbf{x}_{i}$, we define an \textit{extraction candidate set},  $\tilde{\mathbf{x}}_{i}=\{\mathbf{x}_{i}, \mathbf{x}_{i}^{1}, \ldots, \mathbf{x}_{i}^{k}\}$, which includes the document sentence itself and all of its concise versions. For simplicity, we denote the original sentence $\mathbf{x}_{i}$ as $\mathbf{x}_{i}^{0}$, so that $\tilde{\mathbf{x}}_{i}=\{\mathbf{x}_{i}^{0}, \ldots, \mathbf{x}_{i}^{k}\}$. 
Then, we hierarchically encode an extraction candidate in three levels: (1) candidate level, (2) candidate set level, and (3) document level, as illustrated in Figure~\ref{fig:cand-encoder}. 



We first feed each sentence candidate ${\mathbf{x}}_{i}^{j}$ to the Sentence-BERT model~\cite{DBLP:conf/emnlp/Sentence-BERT19} to learn word representations. The Sentence-BERT model incorporates the context information of the sentence candidate into each word representation. It has been shown that the Sentence-BERT model learns better sentence-level representations than the BERT model~\cite{DBLP:conf/naacl/BERT19} for several down-streaming tasks~\cite{DBLP:conf/emnlp/Sentence-BERT19}. We feed the last layer output of Sentence-BERT through a MLP layer to yield the word embeddings. 
We then apply a temporal convolutional neural network~\cite{DBLP:conf/emnlp/Kim14} on the word embeddings of each sentence candidate ${\mathbf{x}}_{i}^{j}$ to learn a \textit{local candidate representation} $\mathbf{r}_{i}^{j}\in\mathbb{R}^{d_{loc}}$. 
To learn a representation for a sentence candidate set $\tilde{\mathbf{x}_{i}}$, we use mean pooling to aggregate the local representations of its candidates $\{\mathbf{r}_{i}^{0},\ldots,\mathbf{r}_{i}^{k}\}$ into a \textit{candidate set representation} $\mathbf{r}_{i}$, i.e., $\mathbf{r}_{i}=\frac{1}{k+1}\sum_{j=0}^{k}\mathbf{r}_{i}^{j}$. 

To incorporate the context information of other candidate sets in the document, we apply a bi-directional LSTM to read all the $n$ candidate set representations $(\mathbf{r}_{1},\ldots,\mathbf{r}_{n})$ and produce a \textit{document-level candidate representation} $\mathbf{h}_{i}\in\mathbb{R}^{d_{doc}}$ for each candidate set $\tilde{\mathbf{x}_{i}}$. 
The \textit{context-aware candidate representation} $\mathbf{c}_{i}^{j}$ of an extraction candidate ${\mathbf{x}}_{i}^{j}$ is then the concatenation of its local representation $\mathbf{r}_{i}^{j}$ and the document-level representation of its associated candidate set $\mathbf{h}_{i}$, i.e., $\mathbf{c}_{i}^{j}=[\mathbf{r}_{i}^{j};\mathbf{h}_{i}]\in\mathbb{R}^{d_{mem}}$. 

\smallskip
\noindent \textbf{Pointer network. }
The pointer network consists of an LSTM decoder, $LSTM_{2}$, and a two-hop attention mechanism. 
To prepare for the attention mechanism, we use all the context-aware candidate representations $\mathbf{c}_{i}^{j}$ to form a memory bank matrix, $\mathbf{B}=[\mathbf{c}_{1}^{0}, \ldots,  \mathbf{c}_{1}^{k}, \ldots, \mathbf{c}_{n}^{0},\ldots, \mathbf{c}_{n}^{k}]$, and use $\mathbf{b}_{l}\in\mathbb{R}^{d_{mem}}$ to denote the $l$-th column vector of $\mathbf{B}$. Thus, the attention mechanism can directly attend to all the context-aware candidate representations. 
For the sake of the later extraction process, we also construct a proxy document, $\bar{\mathbf{X}}=(\mathbf{x}_{1}^{0}, \ldots, \mathbf{x}_{1}^{k}, \ldots, \mathbf{x}_{n}^{0}, \ldots, \mathbf{x}_{n}^{k})$, that aligns each extraction candidate to each column vector of the memory bank. We use $\bar{\mathbf{x}}_{l}$ to denote the $l$-th candidate sentence in $\bar{\mathbf{X}}$. 

At each decoding step $t$, $LSTM_{2}$ first attends its hidden state $\mathbf{z}_{t}\in\mathbb{R}^{d_{dec}}$ to each candidate representation $\mathbf{b}_{l}$ to compute a glimpse vector $\mathbf{e}_{t}\in\mathbb{R}^{d_{mem}}$:
%
\begin{align}
a^{t}_{l} &= \mathbf{v}_{g}^{T} \tanh (\mathbf{W}_{g1} \mathbf{b}_{l} + \mathbf{W}_{g2} \mathbf{z}_{t})\text{,}\\
\alpha^{t}_{l} &= \text{softmax}(a^{t}_{l}) \text{,} \\
\mathbf{e}_{t} &= \sum_{l=1}^{n\times (k+1)} \alpha^{t}_{l} \mathbf{W}_{g1} \mathbf{b}_{l} \text{,}
\end{align}
where $\mathbf{W}_{g1}\in \mathbb{R}^{d_{att}\times d_{mem}}$, $\mathbf{W}_{g2}\in \mathbb{R}^{d_{att}\times d_{dec}}$, and $\mathbf{v}_{g}\in \mathbb{R}^{d_{att}}$ are model parameters. The glimpse vector is then attended to the candidate representations to produce the probability of extracting each sentence candidate $\bar{\mathbf{x}}_{l}$: 
%
\begin{align}
u_{l}^{t} = \mathbf{v}_{p}^{T} \tanh (\mathbf{W}_{p1} \mathbf{b}_{l} + \mathbf{W}_{p2} \mathbf{e}_{t}) \text{,}\\
P_{\theta_{ext}}(l|l_{1},\ldots,l_{t-1}, \bar{\mathbf{X}})=\text{softmax}(u_{l}^{t}) \text{,}
\end{align}
where $\theta_{ext}$ denotes all model parameters of the extractor, $l_{t}$ denotes the index of the candidate extracted at step $t$, $\mathbf{v}_{p}\in \mathbb{R}^{d_{att}}$, and $\mathbf{W}_{p1},\mathbf{W}_{p2}\in \mathbb{R}^{d_{att}\times d_{mem}}$. We simply extract the candidate with the highest extraction probability, and the representation of the extracted candidate $\mathbf{b}_{l_{t}}$ is adopted as the input to $LSTM_{2}$ at the next step. 

\smallskip
\noindent \textbf{Maximum-likelihood pre-training. }
We use the maximum likelihood (ML) loss to pre-train the extractor module. {ML pre-training is a common practice in previous literature~\cite{DBLP:conf/cvpr/RennieMMRG17, DBLP:conf/acl/ChanCWK19, DBLP:conf/acl/BansalC18} to improve the stability of RL training. }
Since we do not have extraction labels for the proxy document, we use the following method to construct those labels.
For each summary sentence $\mathbf{y}_{t}$, we find the most similar sentence candidate from the proxy document in terms of Rouge-L $F_{1}$ score: ${l_{t}}=\text{argmax}_{i} \text{Rouge-L}_{F_{1}}(\bar{\mathbf{x}}_{i}, \mathbf{y}_{t}) \text{.}$
With these extraction labels, we can pre-train an extractor using maximum likelihood loss: $- \sum_{t=1}^{m} \log P_{\theta_{ext}}(l_{t}|l_{1},\ldots, l_{t-1}, \bar{\mathbf{X}})$. 

\subsection{Reinforced Extractor}
We apply reinforcement learning (RL) to fine-tune our extractor module to optimize the ROUGE scores. The process of extracting sentence candidates is formulated as a RL problem as follows. At each time step $t$, the agent selects a sentence candidate $\hat{l}_{t}$ (action) sampled from its policy $\pi_{\theta_{ext}}(\hat{l}_{t}| \hat{\mathbf{l}}_{1:t-1}, \bar{\mathbf{X}})$, where $\hat{\mathbf{l}}_{1:t-1}$ denotes the actions selected by the agent from step $1$ to $t-1$. 
The environment then gives a reward $r_{t}(\hat{l}_{t}, \hat{\mathbf{l}}_{1:t-1},\bar{\mathbf{X}},\mathbf{Y})$ to the agent and transits to the next step $t+1$ with a new state $\hat{s}_{t+1}=(\hat{\mathbf{l}}_{1:t},\bar{\mathbf{X}},\mathbf{Y})$. 
The policy of the agent is the extractor module. 

To allow the agent to determine the number of sentence candidates to extract, we include a stop action to the policy's action space. 
A trainable parameter $\mathbf{c}_{EOE}\in \mathbb{R}^{c}$ is appended to the memory bank $\mathbf{B}$ so that the extractor module can treat it as one of the sentence candidates. Once the agent selects $\mathbf{c}_{EOE}$, we denote the current time step as $T$, the agent receives a reward $r_{T}$, and the whole extraction process terminates. The final output summary will be $\mathbf{\hat{Y}}=(\bar{\mathbf{x}}_{l_{1}},\ldots,\bar{\mathbf{x}}_{l_{T}})$. 

\smallskip
\noindent \textbf{Marginal Reward. }
To allow the agent to be aware of the impact of each selected sentence to the output summary's overall quality, we use the marginal increase in ROUGE-L $F_{1}$ score of the output summary as the reward for selecting a sentence candidate (not including $\mathbf{c}_{EOE}$). More formally, 
$r_{t}=\text{R-L}((\bar{\mathbf{x}}_{\hat{l}_{1}}, \ldots, \bar{\mathbf{x}}_{\hat{l}_{t}}), \mathbf{Y}) - \text{R-L}((\bar{\mathbf{x}}_{\hat{l}_{1}}, \ldots, \bar{\mathbf{x}}_{\hat{l}_{t-1}}), \mathbf{Y})$, for $t=1,\ldots,m$, where R-L denotes ROUGE-L ${F_{1}}$ score. In contrast, the sentence rewriting method~\cite{DBLP:conf/acl/BansalC18} uses the sentence-level Rouge-L $F_{1}$ as the reward of selecting a candidate, which aims at a local optimum of one selection instead of the global optimum of the entire output summary. 
If the agent selects more sentences than the ground-truth summary, we set the reward to be zero for the extraneous selection steps, i.e., $r_{t}=0$ for $t=m+1,\ldots,T-1$. 
This penalizes the agent from extracting too many sentences. When the agent terminates the extraction process by selecting $\mathbf{c}_{EOE}$, we use the ROUGE-1 $F_{1}$ score of all the selected sentences as the reward: $r_{T}=\text{R-1}(\mathbf{\hat{Y}}, \mathbf{Y})$, which considers the bag-of-words information of the output summary. 

We define return to be the total discounted future reward, $R_{t}= \sum_{\tau=0}^{T-t} \gamma^{\tau} r_{t+\tau}$, where $\gamma\in (0,1]$ is a discount factor. The goal of the agent is to maximize the initial return, $R_{1}$. Thus, we define the loss function of the policy as $\mathcal{L}_{rl}(\theta_{ext}) = -\mathbb{E}_{\hat{\mathbf{l}}_{1:T}\sim \pi_{\theta_{ext}}} [R_{1}]$ . To estimate the gradient of this loss function, we apply the advantage actor-critic (A2C) algorithm~\cite{DBLP:conf/icml/MnihBMGLHSK16} due its low variance of gradient estimation. 

\smallskip
\noindent \textbf{Advantage Actor-Critic. }
We define the state value function $V(\hat{s}_{t})$ to be the expectation of return starting from $\hat{s}_{t}$: $V(\hat{s}_{t})=\mathbb{E}[R_{t}|\hat{s}_{t}]$. Then we define the state-action value function $Q(\hat{s}_{t},\hat{l}_{t})$ to be the expectation of return of taking action $\hat{l}_{t}$ at state $\hat{s}_{t}$: $Q(\hat{s}_{t}, \hat{l}_{t})=\mathbb{E}[R_{t}|\hat{s}_{t},\hat{l}_{t}]$. Then the gradient of $\mathcal{L}(\theta_{ext})$ is then formulated as:  
\begin{equation} \label{eq:pg}
\begin{split}
\nabla_{\theta_{ext}} \mathcal{L}(\theta_{ext}) = \mathbb{E}[\nabla_{\theta_{ext}} \log \pi_{\theta_{ext}}(\hat{l}_{t}|\hat{\mathbf{l}}_{1:T}) A(\hat{s}_{t}, \hat{l}_{t})]\text{,}
\end{split}
\end{equation}
where $A(\hat{s}_{t}, \hat{l}_{t})=Q(\hat{s}_{t},\hat{l}_{t})-V(\hat{s}_{t})$ is an advantage function that measures the advantage of taking an action $\hat{l}_{t}$ over other actions at state $\hat{s}_{t}$. 
We use the return $R_{t}$ as an estimation of $Q(\hat{s}_{t},\hat{l}_{t})$ and use another neural network $V_{\theta_{critic}}(\hat{s}_{t})$, called critic, to approximate the value function $V(\hat{s}_{t})$. 
We also train the critic network to minimize the square loss $\mathcal{L}(\theta_{critic}) = (V_{\theta_{critic}}(\hat{s}_{t}) - R_{t})^2$. 
The critic network shares the same structure as the pointer network in Section~\ref{sec:extractor} but the output layer is changed to a regression layer to learn the value function. 

\section{Experiments}

\subsection{Datasets}
We adopt the \textbf{CNN/DailyMail}~\cite{DBLP:conf/nips/HermannKGEKSB15,DBLP:conf/conll/NallapatiZSGX16} corpus as our benchmark dataset. 
Each sample contains a news article and several bullet points that highlight the important information of the article. We use the news article as the document (781 words on average) and the associated bullet points (56 words on average) as the reference summary. 
We use the script provided in Chen et al.~\cite{DBLP:conf/acl/BansalC18} to preprocess the dataset and we do not anonymize the entities.  After preprocessing, we obtain the standard split of 287,113/13,368/11,490 for training, validation, and testing. 

Following previous work~\cite{DBLP:conf/emnlp/ChenGTSZY18, DBLP:conf/acl/BansalC18}, we also conduct experiments on the \textbf{DUC-2002} dataset in a test-only setup to evaluate the generalization ability of different methods. The DUC-2002 dataset contains 567 documents in total. All of the documents are used as test samples. Each sample contains a news article (630 words on average) with two reference summaries (114 words on average). 

We further conduct experiments on the \textbf{Pubmed}~\cite{DBLP:conf/naacl/CohanDKBKCG18} dataset to evaluate the performance of our framework on summarizing a long document. This dataset uses a scientific article as the source document (3,224 words on average) and its abstract as the reference summary (214 words on average). We use the standard split of 119,924/6,633/6,658 for training, validation, and testing. 

\subsection{Baselines and Comparisons}
We compare our framework with the following strong baselines for text summarization: 
\begin{itemize}
    \item {\textbf{NeuSUM}~\cite{DBLP:conf/acl/ZhaoZWYHZ18}: An extractive model that applies a pointer-network~\cite{DBLP:conf/nips/VinyalsFJ15} to jointly score and select the document sentences. }
    \item {\textbf{JECS}~\cite{syntactic_compress_19}: A hybrid method that follows the select-then-compress framework. It compresses each selected sentence by removing words from the sentence. }
    \item {\textbf{Pointer-Gen. + Cov}~\cite{DBLP:conf/acl/SeeLM17}: The pointer-generator network with the coverage mechanism. }
    \item {\textbf{Deep-Reinforce}~\cite{DBLP:conf/iclr/PaulusXS18}: An abstractive method that augments the pointer-generator network with an intra-attention mechanism and proposes a mixed training objective (RL+ML) to train the model. }
    \item {\textbf{Sentence Rewriting}~\cite{DBLP:conf/acl/BansalC18}: A hybrid method that follows the select-then-compress framework. It uses the pointer-generator network to condense each of the selected sentences. }
    \item {\textbf{SENECA}~\cite{DBLP:conf/emnlp/SENECA19}: A hybrid method that follows the select-then-compress framework. It explicitly incorporates the entity information into the extractor. }
    \item {\textbf{MATCHSUM}~\cite{DBLP:conf/acl/TextMatching20}: An extractive model that performs extraction at the summary level by enumerating all possible combinations of sentences from a pruned document and fine-tune a BERT-based model to rank the summary candidates. 
    }
    \item {\textbf{SAGCopy}~\cite{DBLP:conf/acl/SelfAttnGuidedCopy20}: An abstractive model that utilizes the self-attention distribution to guide the copying process in a large pre-trained language model. }
    \item {\textbf{PEGASUS}~\cite{DBLP:conf/icml/PEGASUS}: A large-scale pre-trained abstractive summarization model. It introduces a gap-sentences generation objective during pre-training. }
\end{itemize}
    {We also compare our framework with the following baselines for long document summarization:}
\begin{itemize}
    \item {\textbf{Discourse-aware}~\cite{DBLP:conf/naacl/CohanDKBKCG18}: An abstractive summarization model for a long document. It uses a hierarchical encoder to capture the discourse structure of the document and a discourse-aware decoder to generate a summary. }
    \item {\textbf{Global-local Attentive}~\cite{DBLP:conf/emnlp/XiaoLongText19}: An extractive summarization model for a long document. It combines the global context representation of the document and the local context representation of the current section/topic using an attention mechanism. }
    \item {\textbf{Global-local Concat}~\cite{DBLP:conf/emnlp/XiaoLongText19}: An extractive summarization model for a long document. It is similar to Global-local Attentive but it uses a concatenation operation to combine the global and local context representations. }
    \item {\textbf{Dancer}~\cite{DBLP:journals/taslp/Dancer20}: An abstractive summarization model for a long document. It divides a document into different parts and uses an abstractor to summarize each part of the document separately. }
    \item {\textbf{Sent-PTR}~\cite{DBLP:conf/emnlp/PilaultLSP20}: An extractive summarization model for a long document. It uses a hierarchical LSTM~\cite{DBLP:journals/neco/HochreiterS97} to encode the document sentences and uses a pointer network~\cite{DBLP:conf/nips/VinyalsFJ15} to extract sentences. }
    \item {\textbf{Extr-Abst-TLM}~\cite{DBLP:conf/emnlp/PilaultLSP20}: A hybrid summarization model for a long document. It uses the Sent-PTR model to encode document sentences then uses a Transformer language model to condense the sentences. }
    \item {\textbf{BIGBIRD-PEGASUS}~\cite{DBLP:conf/nips/bigbird20}: A large-scale pre-trained abstractive summarization model for a long document. It introduces a sparse attention mechanism to reduce the dependency on the sequence length. 
    }
\end{itemize}


The results of our framework with different abstractors are reported. We use \textbf{one2one-long-short}, \textbf{compress-ctrl}, and \textbf{two-to-one} to denote the results of our framework using the one-to-one long-short abstractor, compression-controllable abstractor, and two-to-one abstractor respectively. We use \textbf{one2one-top$\mathbf{k}$} to denote the results of the one-to-one top-$k$ abstractor, e.g., \textbf{one2one-top2} indicates that we take the top-$2$ output sequences from the beam search algorithm as the generated concise versions. 

We evaluate the performance of different summarization methods using the full-length \textbf{ROUGE-1}, \textbf{ROUGE-2}, and \textbf{ROUGE-L} $F_{1}$ scores with stemming~\cite{lin2004rouge}. For brevity, we refer to ROUGE $F_{1}$ scores as ROUGE scores. 
We use the official Perl script to compute ROUGE scores in our experiments. 
{
For the CNN/DM and DUC-2002 datasets, we specify the parameters\footnote{{-c 95 -r 1000 -n 2 -m}} according to Chen and Bansal~\cite{DBLP:conf/acl/BansalC18}. For the Pubmed dataset, we specify the parameters\footnote{{-a -n 2 -r 1000 -f A}} according to Xiao et al.~\cite{DBLP:conf/emnlp/XiaoLongText19}. 
}

\subsection{Implementation Details}
{All our methods are trained and tested on a Nvidia Tesla T4 GPU. }
We use the ``bert-base-nli-mean-tokens'' version of the Sentence-BERT~\cite{DBLP:conf/emnlp/Sentence-BERT19}. The hidden sizes of bi-directional LSTM and uni-directional LSTM are set to 512 and 256 respectively. The initial hidden state for an extractor is a trainable parameter, while we use a linear layer to map the final encoder hidden state to the initial decoder state for an abstractor. The dimension of the latent variable $\beta_{t}$ in our compression-controllable abstractor is 128. The convolution filters in the temporal CNN encoder~\cite{DBLP:conf/emnlp/Kim14} have a hidden size of 100 with windows sizes of 3, 4, and 5. {For the advantage actor-critic algorithm, we follow the hyper-parameter setting in \cite{DBLP:conf/acl/BansalC18}.} The discounted factor $\gamma$ is set to $0.95$. During training, we truncate every article sentence to 100 tokens and summary sentences to 30 tokens. The batch size is set to 32. We apply gradient clipping of 2.0 using L-2 norm. We use the Adam optimization algorithm~\cite{Kingma2014Adam} with an initial learning rate of 5e-4 for ML training and an initial learning rate of 5e-5 for RL training. 
We adopt the diverse beam search algorithm~\cite{DBLP:journals/corr/LiMJ16} as the decoding algorithm for an abstractor. Our source code will be released in the future. 

\begin{table}[t]
\centering
\small
\begin{tabular}{l|cccc}
\hline \hline
\textbf{Model}         & \textbf{Type}      & \textbf{R-1}   & \textbf{R-2}   & \textbf{R-L} \\
\hline \hline
NeuSUM                 & Ext.       & 41.59     & 19.01     & 37.98     \\
{MATCHSUM}               & {Ext.}       & {\textbf{\underline{44.41}}}     & {\underline{20.86}}     & {\underline{40.55}}     \\
Pointer-Gen. + Cov.    & Abs.       & 39.53     & 17.28     & 36.38     \\
Deep-Reinforce$^{\dagger}$ & Abs.   & 39.87     & 15.82     & 36.90     \\
Bottom-Up              & Abs.       & 41.22     & 18.68     & 38.34     \\
DCA                    & Abs.       & 40.91     & 19.21     & 38.03     \\
{SAGCopy}                & {Abs.}       & {42.53}     & {19.92}     & {39.44}     \\
{PEGASUS}         & {Abs.}       & {\underline{44.17}}     & {\textbf{\underline{21.47}}}   &  {\textbf{\underline{41.11}}}     \\
Sentence Rewriting     & Hyb.       & 41.07     & 17.96     & 38.59     \\
JECS                   & Hyb.       & 41.70     & 18.50     & 37.90     \\
SENECA                 & Hyb.       & 41.52     & 18.36     & 38.09     \\  \hline
\hline
{\bf Our Models}             &                 &    &    &  \\
One2one-top$1$ w/o S.BERT      & Hyb.       & 42.17$^{*}$    & 19.36     & 38.97$^{*}$  \\
Compression-ctrl w/o S.BERT    & Hyb.       & 42.23$^{*}$    & 19.30     & 39.13$^{*}$    \\
One2one-top$1$                 & Hyb.       & 42.51$^{*}$    & 19.52$^{*}$     & 39.13$^{*}$    \\
Compression-ctrl               & Hyb.       & \underline{42.71}$^{*}$   & \underline{19.59}$^{*}$     &  \underline{39.34}$^{*}$    \\
\hline
\end{tabular}
\caption{
Comparison results on the CNN/DM dataset. {We underline the highest scores in each category and bold the overall highest scores. }
The suffix ``w/o S.BERT'' denotes that we replace the Sentence-BERT word embeddings~\cite{DBLP:conf/emnlp/Sentence-BERT19} with the word2vec word embeddings~\cite{DBLP:conf/nips/MikolovSCCD13}. 
$^{*}$ indicates results that are significantly better than the baselines ($p<0.03$, t-test) {except MATCHSUM, SAGCopy, and PEGASUS}. 
$^{\dagger}$ indicates significant test done on outputs by our implementation, which achieves comparable ROUGE-1,2,L scores (40.25, 17.11, 37.57). 
}
\label{table:main-results}
\vspace{-0.1in}
\end{table}


\begin{table}[t]
\centering
\small
 \setlength{\tabcolsep}{0.8mm}
\begin{tabular}{l|ccc|cc|c}
\hline \hline
\textbf{Model}                                                       & \textbf{R-1 recall}   & \textbf{R-2 recall}  & \textbf{R-L recall} & \textbf{Org. \%}   & \textbf{Con. \%} & {\textbf{Avg. len.}} \\
\hline \hline
Sent. Rewrit.          & 45.56     &    19.92  & 42.79  & 0.0    & 100.0 &  {68.9} \\
\hline
One2one-top$1$                                     & 49.12     & 22.63  & 45.20  & 49.3   &  50.7 & {70.8} \\
Compress-ctrl                                      & \textbf{49.39}     & \textbf{22.69}  &   \textbf{45.47}  & 45.5    & 54.5 & {69.2} \\
\hline
\end{tabular}
\caption{ROUGE recall scores on the CNN/DM dataset. Sent. Rewrit. denotes the sentence rewriting method. 
Org. \% denotes the percentage of selected sentences that are original document sentences. Con. \% denotes the percentage of selected sentences that are condensed document sentences. 
Avg. len. denotes the average number of words per summary. 
}
\label{table:recall}
\end{table}

\subsection{Main Results}
We present our main results\footnote{We re-evaluate the results of DCA~\cite{DBLP:conf/naacl/CelikyilmazBHC18} using full-length summary since they reported an evaluation setting using truncated summary.} on the CNN/DM dataset in Table~\ref{table:main-results}. 
We have the following observations:
\begin{itemize}
    \item \textit{Our condense-then-select framework outperforms strong baselines.} 
    Without using the Sentence-BERT embeddings, both of the one-to-one top-$1$ abstractor and compression-controllable abstractor of our framework achieve significantly higher ROUGE scores than the baselines except MATCHSUM, SAGCopy, and PEGASUS. These results demonstrate the effectiveness of the condense-then-select paradigm in our framework. 
    {The PEGASUS and SAGCopy models fine tune a large pre-trained language model which has more than 120M parameters. On the other hand, our methods have a substantially fewer number of parameters to train. Our largest model (compress-ctrl) only needs to train \~{}9M parameters since we freeze the parameters of the sentence-BERT model. Hence, the PEGASUS and SAGCopy models can fit the distribution of the CNN/DM dataset better and achieve higher ROUGE scores than our methods. 
    The MATCHSUM model enumerates all possible combinations of sentences in a pruned document as summary candidates. Then it fine-tunes a BERT-based model to rank the summary candidates. In total, this model needs to rank $\binom{n'}{m}$ summaries, where $n'$ is the number of sentences in the pruned document and $m$ is the number of summary sentences. In contrast, the extractor in our methods compose a summary in $m$ extraction steps. Our extractor ranks $\tilde{n}$ candidate sentences in each extraction step and thus only needs to rank $\tilde{n} \times m$ sentences in total. This suggests a trade-off between extractor performance and time complexity. 
    }
    \item {\textit{Compression-controllable abstractor outperforms one-to-one top-$1$ abstractor}. It is because our compression-controllable abstractor provides concise versions with different compression levels, which allow the extractor to achieve a better balance between information coverage and brevity of the output summary. }
    \item \textit{The Sentence-BERT embeddings further improve the performance of our methods.} After applying the Sentence-BERT embeddings, both of the one-to-one top-$1$ abstractor and compression-controllable abstractor of our framework obtain higher ROUGE scores. It is because the Sentence-BERT model is pre-trained on huge text corpora and is fine-tuned on a sentence-level natural language inference dataset. Hence, the Sentence-BERT model can learn meaningful contextual embeddings for the words in a sentence candidate. These embeddings help the extractor identify salient information from the extraction candidates. 
    \item {\textit{The Sentence-BERT embeddings increase the training time of our extractor.}
    Without using the Sentence-BERT embeddings, the extractor in our framework takes around 6 hours for ML training and 12 hours for RL training on a single Nvidia T4 GPU. After applying the Sentence-BERT embeddings, our extractor takes 1.5 days for ML training and 2.5 days for RL training in the same environment. }
\end{itemize}





\subsection{Preservation of Salient Information}
{In this section, we verify whether our framework can preserve more salient information than the sentence rewriting method. We use ROUGE recall as a proxy metric to measure the amount of salient information in the predicted summaries. }
Table~\ref{table:recall} shows the ROUGE recall scores obtained by our framework and the sentence rewriting method~\cite{DBLP:conf/acl/BansalC18}. 
Both methods of our framework achieve significantly higher ROUGE recall scores than the sentence rewriting method, demonstrating that more salient information is retained in the generated summaries. 

To understand the reasons behind the ROUGE recall improvement, we also report the percentage of selected sentences that are original document sentences and the percentage of selected sentences that are condensed document sentences in Table~\ref{table:recall}. 
{To compute the condense \%, we count the total number of extracted sentence candidates that belong to original document sentences in all the output summaries, and then divide it by the total number of extracted sentence candidates in all the output summaries. The original \% is computed in a similar way.} We observe that a significant proportion of the sentences selected by our framework belongs to original document sentences, suggesting that our framework selects original sentences when appropriate to retain more salient information in the output summaries. 
{Since our methods extract both original document sentences and condensed sentences, our predicted summaries have a longer average length compared to the sentence rewriting method, as shown in Table~\ref{table:recall}. }



\begin{table}[t]
\centering
\small
\begin{tabular}{l|ccc}
\hline \hline
\textbf{Model}                                                       & \textbf{R-1}   & \textbf{R-2}   & \textbf{R-L} \\
\hline \hline
One2one-top$2$                                            & 42.33     & 19.41     &  39.01    \\
One2one-long-short                                      &  42.29    & 19.21     &  38.98    \\
Compress-ctrl                                     & \textbf{42.71}   & \textbf{19.59}     &  \textbf{39.34} \\
two-to-one                                                    &  41.42    & 18.71     &  37.83    \\
\hline
{Compress-ctrl + top$2$}                                      & {42.34}     & {19.20}     & {38.86}     \\
{two-to-one + top$2$}                                       & {41.68}     & {18.84}     & {38.07}     \\
\hline
\end{tabular}
\caption{Results of our condense-then-select approach with different abstractors on CNN/DM. 
{
The top segment shows the abstractors that generate two concise versions for each document sentence. 
The bottom segment shows the combination of candidates from two abstractors.
}
}
\label{table:beam-analysis}
\end{table}

\subsection{Performance of Different Abstractors}
We compare the performance of different abstractors that generate two concise versions for a document sentence. The results are shown in the top segment of Table~\ref{table:beam-analysis}. We observe that our compression-controllable abstractor outperforms all other abstractors. Different from other one-to-one abstractors, which generate two concise versions by simply taking two sequences from the beam search, our compression-controllable abstractor is explicitly trained to produce two concise versions with different compression levels. Thus, the generated candidates have higher qualities. We also observe that the results of two-to-one abstractor are significantly worse than the others. We investigate whether a secondary source sentence in our two-to-one abstractor can provide useful information for generating a summary. First, we use Eq.~(\ref{eq:primary-sentence}) and Eq.~(\ref{eq:secondary-sentence}) to align each ground-truth summary sentence $\mathbf{y}_{t}$ to a source sentence $\mathbf{x}_{j_t}$ and a secondary source sentence $\mathbf{x}_{j'_t}$. Then we compute the improvement in ROUGE-L recall score to each summary sentence brought by its secondary source sentence, $\text{Rouge-L}_{r}([\mathbf{x}_{j_t}; \mathbf{x}_{j'_t}], \mathbf{y}_{t}) - \text{Rouge-L}_{r}(\mathbf{x}_{j_t}, \mathbf{y}_{t})$. Only 16.7\% of the secondary source sentences lead to an improvement of more than 25 on ROUGE-L recall, demonstrating that most of the secondary source sentences can only provide little information about the summary sentence. Thus, the abstractor has a heavier burden of handling redundant and unrelated content in the input, leading to a decrease in performance. 

{
Moreover, we verify whether combining the condensed candidates from different abstractors will lead to better performance. We try two different combinations: combination of candidates from compress-ctrl and one2one-top$2$ abstractors (compress-ctrl + top$2$); and combination of candidates from two-to-one and one2one-top$2$ abstractors (two-to-one + top$2$). The results are reported in the bottom segment of Table~\ref{table:beam-analysis}. We find that combining the candidates from different abstractors yield lower ROUGE scores. We suspect that the increase in action space makes the extractor more difficult to learn to select appropriate sentence candidates. Hence, we conduct an empirical study about the effect of candidate size on the summarization performance in the next section. 
}

\begin{table}[t]
\centering
\small
\begin{tabular}{l|ccc}
\hline \hline
\textbf{Model}                                                       & \textbf{R-1}   & \textbf{R-2}   & \textbf{R-L} \\
\hline \hline
One2one-top$1$                                            & \textbf{42.51}    & \textbf{19.52}     & \textbf{39.13}    \\
One2one-top$2$                                            & 42.33     & 19.41     &  39.01    \\
One2one-top$3$                                            & 41.92     & 19.03     &  38.60     \\
\hline
\end{tabular}
\caption{Results of the condense-then-select framework with one-to-one abstractor using different numbers of extraction candidates.}
\label{table:action-space}
\end{table}

\subsection{Effects of Candidate Size} 
We investigate whether we can improve the performance of our framework by simply increasing the number of extraction candidates. We compare the performance of one2one-top$1$, one2one-top$2$, and one2one-top$3$ abstractors. These methods provide different numbers of extraction candidates for our framework. The results are shown in Table~\ref{table:action-space}.
As can be seen, the performance of our framework drops when the number of condensed candidates increases. It is because by enlarging the number of candidates per input sentence, the action space of the agent increases\footnote{Though RL suffers from the problem of exploration, it still significantly improves the performance of our methods. Our framework with one2one-top2 abstractor without RL training only obtains ROUGE of (39.45, 17.12, 36.96).}. If the newly added candidates are of low quality, the performance of the extractor will decrease. Indeed, we do observe the quality of candidates deteriorates in informativeness and fluency. This suggests a future direction when diverse candidates of high quality should be produced. 


\subsection{Cross-domain Analysis}
Similar to Chen and Bansal~\cite{DBLP:conf/acl/BansalC18}, we conduct experiments on the test-only DUC-2002 dataset to evaluate our methods on out-of-domain data. All the methods are trained on the CNN/DM dataset and then tested on the DUC-2002 dataset. 
The results are shown in Table~\ref{table:cross-domain}. 
{First, we observe that hybrid methods have a strong generalization ability. 
Our methods achieve substantially higher ROUGE scores than all the abstractive baselines; 
while the sentence rewriting method outperforms all the abstractive baselines except PEGASUS. 
}
The reason is that hybrid methods explicitly divide the text summarization process into sentence selection and sentence abstraction. Hence, they impose a strong inductive bias on the neural network models, which help the models generalize to out-of-domain data. In contrast, the abstractive baselines use an encoder-decoder model to perform the entire text summarization process. Hence, these abstractive methods have a weaker inductive bias. 
Second, we observe that our methods outperform the sentence-rewriting method. It is because the select-then-compress framework suffers from the problem of loss of salient information, while our condense-then-select framework retains more salient information by including original document sentences in the extraction candidates. 

\begin{table}[t]
\centering
\small
\begin{tabular}{l|cccc}
\hline \hline
\textbf{Model}                       & \textbf{Type}   & \textbf{R-1}   & \textbf{R-2}   & \textbf{R-L} \\
\hline \hline
{MATCHSUM}                             & Ext.   & 43.01 & 23.86 & 39.58 \\
Pointer-Gen + Cov.                   & Abs.   & 37.22 & 15.78 & 33.90 \\
RL + Intra.Attn                      & Abs.   & 36.49 & 16.62 & 34.68 \\
Bottom-up                            & Abs.   & 34.70 & 17.33 & 32.50 \\
{PEGASUS$^{\ddagger}$}           & {Abs.}   & {38.35} & {19.95} & {35.62} \\ 
Sentence Rewriting                   & Hyb.   & 40.12 & 19.57 & 37.63 \\ 
{\bf Our Models}                     &  &    &    &  \\
One2one-top$1$                & Hyb.   & \textbf{43.94} & \textbf{24.20}  & \textbf{40.83} \\
Compress-ctrl                 & Hyb.   & 42.92 & 23.05 & 40.04 \\
\hline
\end{tabular}
\caption{Results on the test-only DUC-2002 dataset. 
{$^{\ddagger}$: the outputs are decoded from the ``PEGASUS-CNN\_DailyMail'' checkpoint provided by the HuggingFace Transformers~\cite{wolf-etal-2020-transformers}. }
}
\label{table:cross-domain}
\end{table}

\begin{table}[t]
\centering
\small
\begin{tabular}{l|cccc}
\hline \hline
{\textbf{Model}}                  & {\textbf{Type}}   & {\textbf{R-1}}   & {\textbf{R-2}}   & {\textbf{R-L}} \\
\hline \hline
{MATCHSUM}                   & Ext.   & 41.21 & 14.91 & 36.75 \\
{Global-local Attentive}          & Ext.   & 44.81 & \underline{19.74} & 31.48 \\
{Global-local Concat}             & Ext.   & \underline{44.85} & 19.70 & 31.43 \\
{Sent-PTR}                        & Ext.   & 43.30 & 17.92 & \underline{39.47} \\
{Discourse-aware}                 & Abs.   & 38.93 & 15.37 & 35.21 \\ 
{PEGASUS}                    & Abs.   & {45.97} & {20.15} & {41.34} \\
{BIGBIRD-PEGASUS}                 & Abs.   & \textbf{46.32} & \underline{20.65} & \textbf{42.33} \\
{Extr-Abst-TLM}                   & Hyb.   & 42.13 & 16.27 & 39.21 \\
{Dancer}                          & Hyb.   & 44.09 & 17.69 & 40.27 \\
{Sentence Rewriting}              & Hyb.   & 44.68 & 19.07 & 41.33 \\
\hline
{\bf Our Models}                &    &    &   &   \\
{One2one-top$1$ w/o S.BERT}       & Hyb.   & 46.00    & 20.87     & 41.48    \\
{Compression-ctrl w/o S.BERT}     & Hyb.   & \underline{46.10}    & \textbf{20.92}     & \underline{41.64}    \\
\hline
\end{tabular}
\caption{{Results on the Pubmed dataset. We underline the highest scores in each category and bold the overall highest scores. The suffix ``w/o S.BERT'' denotes that we replace the Sentence-BERT word embeddings with the word2vec word embeddings. }
}
\label{table:pubmed-results}
\end{table}

\subsection{{Results on Long Document Summarization}}
{We further conduct experiments on a long document summarization dataset, Pubmed, and the results are presented in Table~\ref{table:pubmed-results}. We have the following observations:}
\begin{itemize}
    \item {\textit{Our condense-then-select framework outperforms strong long text summarization baselines.} 
    We can see that our methods obtain substantially higher ROUGE scores than all the long text summarization baselines except the BIGBIRD-PEGASUS model. Moreover, the performance of our methods is competitive to the BIGBIRD-PEGASUS model which contains more than 300M parameters. 
    These results demonstrate the effectiveness of our condense-then-extract framework on summarizing a long text document. 
    }
    \item {\textit{The training speed of our methods on the Pubmed dataset becomes very slow when using the Sentence-BERT embeddings. } 
    We observe that our extractor takes around 4 days for ML training on a Nvidia T4 GPU. For RL training, our extractor has not converged after training for 6 days in the same environment, which means it will take more than 10 days to train our extractor on the Pubmed dataset. 
    Hence, we do not use the Sentence-BERT embeddings on the Pubmed dataset due to the long training time. Without applying the Sentence-BERT embeddings, our extractor only takes around 12 hours for ML training and 2 days for RL training in the same environment. 
    }
\end{itemize}

\subsection{Human Evaluation}
We conduct human evaluation to assess the qualities of the summaries generated by our methods. We use the following three evaluation criteria: (1) \emph{informativeness} measures the amount of salient information in a summary; (2) \emph{conciseness} measures how well a summary avoids redundant information; and (3) \emph{readability} measures the grammaticality and fluency. 
We hire three student helpers as annotators to evaluate 50 randomly selected test samples from the CNN/DM dataset. 
For each test sample, we show the document, the ground-truth summary, and the summaries generated by our framework with the compression-controllable abstractor, one2one-top1 abstractor, and the sentence rewriting method~\cite{DBLP:conf/acl/BansalC18}. The generated summaries are anonymized and randomly shuffled. Every generated summary is rated by all the annotators, and the rating scale is from 1 to 5. The complete human evaluation guideline in the Appendix. 
We report the averaged scores for each method in Table~\ref{table:human}. 
Results show that both of our methods obtain substantially higher informativeness, conciseness, and readability scores than the sentence rewriting method. 

\begin{table}[t]
\centering
\small
\begin{tabular}{l|ccc}
\hline \hline
\textbf{Model}                                                       & \textbf{Informativeness} & \textbf{Conciseness} & \textbf{Readability} \\
\hline \hline
Sentence Rewriting                                         & 2.99   & 3.08   & 3.97   \\
One2one-top1                                               & 3.15   & 3.29   & 4.10   \\
Compress-ctrl                                              & \textbf{3.35}   & \textbf{3.44}   & \textbf{4.21}   \\
\hline
\end{tabular}
\caption{
Human evaluation scores of informativeness, conciseness, and readability on 50 random test samples of the CNN/DM dataset. 
}
\label{table:human}
\end{table}


\begin{figure}[t]
\centering
\small
\fontsize{10}{8}\selectfont
\begin{tabular}{|p{0.95\columnwidth}|}
\hline
\textbf{Ground-truth:} \red{sulforaphane known to block inflammation and damage to the cartilage .} \orange{people would have to eat several pounds daily to derive significant benefit .} \green{drug company evgen pharma has developed synthetic version of chemical .} \\  \hline
\textbf{Compress-ctrl:} \red{the broccoli chemical sulforaphane is known to block the inflammation and damage to cartilage associated with the condition .} \orange{patients would have to eat several pounds of the vegetable every day to derive any significant benefit .} \green{evgen pharma developed a stable synthetic version of the chemical .} \\ \hline
\textbf{One2one-top1:} \red{the broccoli chemical sulforaphane is known to block the inflammation and damage to cartilage associated with the condition .}
\orange{patients would have to eat several pounds of vegetable every day to derive any significant benefit .} \green{evgen pharma has developed a stable synthetic version of the chemical that offers potential of a pill treatment .} \\ \hline
\textbf{Sentence rewriting:} \red{the broccoli chemical sulforaphane is known to block inflammation and damage to cartilage associated with arthritis . broccoli chemical sulforaphane is known to block the inflammation and damage to cartilage .} \green{uk drug company evgen pharma has developed a stable synthetic version .} \orange{patients would have to eat several pounds of the vegetable every day .} \\ \hline
\end{tabular}
\caption{A case study compares the summaries generated by different methods on the CNN/DM testing set. 
Different colors are used to highlight the information in each ground-truth summary sentence. }
\vspace{-0.1in}
\label{figure:case-study}
\end{figure}

\subsection{Case Study}
We compare the output summaries of the above three methods in Figure~\ref{figure:case-study}. 
Different colors are used to highlight the information in each ground-truth summary sentence. 
We make the following observations: 
\begin{itemize}
    \item {\textit{The fourth sentence predicted by the sentence rewriting method deletes the salient information (``to derive significant benefit'') that appears in the ground-truth, while both of our methods retain that information.} It is because the extractor in our methods can select an original document sentence when salient information is removed during sentence compression. Hence, our methods achieve a higher informativeness score in Table~\ref{table:human} and higher ROUGE scores in Table~\ref{table:main-results} compared to the sentence rewriting method. } 
    \item {\textit{The third sentence predicted by our method with one2one-top1 abstractor includes the redundant information of ``that offers potential of a pill treatment'', while our method with compression-controllable abstractor avoids such redundant information.} It is because the compression-controllable abstractor generates condensed sentences with different compression levels, which allow the extractor to select a more concise sentence candidate that covers the salient information. Therefore, our method with compression-controllable abstractor obtains a higher conciseness score and higher ROUGE scores than our method with one-to-one top-$1$ abstractor. }
\end{itemize}

{
From the output summaries, we also observe that the sentence rewriting method is likely to delete the words after a preposition and the words after a conjunction. 
We illustrate more samples of output sentences in Figure~\ref{figure:omit-words-samples}. 
By comparing the ground-truth summary sentences and the sentences generated by the sentence rewriting method, we can see that the sentence rewriting method omits the words after the preposition ``to'' and the words after the conjunction ``after''. 
These samples suggest that the sentence-level abstractor may exploit some simple rules to condense a sentence. 
On the other hand, our Compress-ctrl method retains those words omitted by the baseline. It is because the extractor in our framework can utilize the context information of the document to determine whether to select an original document sentence or a condensed sentence. 
}

\begin{figure}[t]
\centering
\small
\fontsize{10}{8}\selectfont
\begin{tabular}{|p{0.95\columnwidth}|}
\hline
\textbf{Example 1}\\
\hline
\textbf{Ground-truth:} she says she will be using a voice phone \red{to dictate her tweets} .\\  \hline
\textbf{Sentence rewriting:} the prisoner formerly known as bradley manning said she will be using a voice phone .\\ \hline
\textbf{Compress-ctrl:}  the prisoner formerly known as bradley manning said she will be using a voice phone \red{to dictate her tweets} .\\ \hline
\textbf{Example 2}\\
\hline
\textbf{Ground-truth:} james oliver , 48 , was left with a serious leg injury \red{after being allegedly hit by a car driven by linda currier , 53} .\\  \hline
\textbf{Sentence rewriting:} james oliver , 48 , was left with a serious leg injury .\\ \hline
\textbf{Compress-ctrl:} james oliver , 48 , was left with a serious leg injury \red{after being allegedly hit by a car driven by linda currier , 53} .\\ \hline
\end{tabular}
\caption{
{Sample summary sentences generated by different methods on the CNN/DM testing set. 
The words omitted by the sentence rewriting method are highlighted.}
} 
\vspace{-0.1in}
\label{figure:omit-words-samples}
\end{figure}

\section{Conclusion}
We propose a novel condense-then-select framework that first condenses document sentences and then performs sentence selection to assemble a summary. By allowing the extractor to select among the original document sentences and their concise versions, our framework outperforms the select-then-compress framework and other strong baselines on the CNN/DM dataset. 
Cross-domain analysis on the DUC-2002 dataset shows that our framework achieves better generalization than the baselines. 
{The experiment results on the Pubmed dataset demonstrate the performance of our framework on summarizing a long text document. }
We also investigate different types of abstractors within our framework. The most effective one is our compression-controllable abstractor. This work provides an interesting research direction, where one could propose and apply different abstractors to generate diverse extraction candidates. 

\section*{Acknowledgements}
The work described in this paper was partially supported by the Research Grants Council of the Hong Kong Special Administrative Region, China (CUHK 2410021, Research Impact Fund, R5034-18), National Key Research and Development Program of China (No. 2018AAA0100204), Amazon AWS Machine Learning Research Award (MLRA) 2020, the Science and Technology Development Fund of Macau SAR (File no. 0015/2019/AKP), Guangdong-Hong Kong-Macao Joint Laboratory of Human-Machine Intelligence-Synergy Systems (No. 2019B121205007), and the National Natural Science Foundation of China (No. 61803083). We would like to thank Prof. Lu Wang from University of Michigan for the helpful suggestions. We would also like to thank the editors and the three anonymous reviewers for their comments. 

\appendix
\section{{Human Evaluation Guidelines}}
{
The annotators are asked to evaluate each summary on the following three aspects: 
\begin{enumerate}
    \item \textbf{Informativeness}: measures how well a summary retains the salient information of the document. Each annotator gives a rating from 1 to 5. 5 indicates the summary retains all the key information of the document. 4 indicates the summary covers most of the key information of the document. 3 indicates that several key points of the document are missing, but it is still in general relevant. 2 indicates the summary only contains a little information of the document. 1 indicates that the summary does not contain any information of the document. 
    \item \textbf{Conciseness}: measures how well a summary avoids redundant information. Each annotator gives a rating from 1 to 5. 5 indicates the summary does not contain any redundant information. 4 indicates the summary contains little redundant information. 3 indicates the summary contains several pieces of redundant information. 2 indicates that the amount of redundant information is larger than the salient information in the summary. 1 indicates the summary is extremely redundant. 
    \item \textbf{Readability}: measures the readability of a summary and how well a summary avoids grammatical errors. Each annotator gives a rating from 1 to 5. 5 indicates the summary has no grammatical error. 4 indicates the summary has rare and minor grammatical errors. 3 indicates that the summary has several grammatical errors but it is in general readable. 2 indicates that the summary has unreadable fragments, but it still has fluent segments that are readable. 1 indicates the summary is completely unreadable. 
\end{enumerate}
}

\bibliography{mybibfile}

\end{document}